\documentclass[]{spie}  

 \usepackage[]{graphicx}
\usepackage{subcaption}
\usepackage{amsmath,amsfonts,amssymb}
\usepackage{graphicx}
\usepackage[colorlinks=true, allcolors=blue]{hyperref}

\title{Automatic Microscopic Cell Counting by Use of Unsupervised Adversarial Domain Adaptation and Supervised Density Regression}
\author[1]{Shenghua He}
\author[2,3]{Kyaw Thu Minn}
\author[3]{Lilianna Solnica-Krezel}
\author[4]{Hua Li}
\author[1,2,4]{Mark Anastasio}
\affil[1]{Department of Computer Science and Engineering}
\affil[2]{Department of Biomedical Engineering}
\affil[3]{Department of Developmental Biology}
\affil[4]{Department of Radiation Oncology\break Washington University in St$.\,$Louis, St$.\,$Louis, MO, USA}

\authorinfo{Further author information: (Send correspondence to Hua Li \& Mark Anastasio)\\E-mail: \{li.hua, anastasio\}@wustl.edu, Telephone: 1 314 537 7145 \& 1 314 935 3637}

\begin{document} 
\maketitle

\begin{abstract}
Accurate cell counting in microscopic images is important for medical diagnoses and biological studies.
However, manual cell counting is very time-consuming, tedious, and prone to subjective errors. 
We propose a new density regression-based method for automatic cell counting that reduces the need to manually annotate experimental images.
A supervised learning-based density regression model (DRM) is trained with annotated synthetic images (the source domain) and their corresponding ground truth density maps.
A domain adaptation model (DAM) is built to map experimental images (the target domain) to the feature space of the source domain.
By use of the unsupervised learning-based DAM and supervised learning-based DRM, a cell density map of a given target image can be estimated, from which the number of cells can be counted.
Results from experimental immunofluorescent microscopic images of human embryonic stem cells demonstrate the promising performance of the proposed counting method.

\end{abstract}

\keywords{Automatic cell counting, microscopic images, density regression, domain adaptation, supervised learning, unsupervised adversarial learning}

\section{INTRODUCTION}
\label{sec:intro}  

Accurately counting the number of cells in microscopic images is important for medical diagnoses and biological applications~\cite{venkatalakshmi2013automatic}.
However, manual cell counting is a very time-consuming, tedious, and error-prone task. An automatic and efficient solution to improve the counting accuracy is highly desirable, but automatic cell counting is challenged by the low image contrast and significant inter-cell occlusions in 2D microscopic images~\cite{matas2004robust,barinova2012,arteta2012,xing2014automatic}.

Recently, density regression-based counting methods
have been employed to address these challenges~\cite{lempitsky2010,xie2018}.
These methods employ machine-learning tools to learn a density regression model (DRM) that estimates the cell density distribution from the characteristics/features of a given image.
The number of cells can be subsequently counted by integrating the estimated density map.
However, in these methods, a large amount of annotated data are required, which can be difficult to obtain in practice.
Using annotated synthetic images to train a DRM can mitigate this problem~\cite{xie2018}.
However, due to the intrinsic domain shift between the experimental and synthetic cell images, a DRM trained with synthetic images may generalize poorly to experimental images~\cite{lempitsky2010,xie2018}.
Currently available DRM methods~\cite{lempitsky2010,xie2018} cannot address this critical issue.

Domain adaptation methods based on unsupervised adversarial learning have been proposed 
to mitigate the harmful effects of domain shift in computer vision tasks, such as classification~\cite{tzeng2017adversarial} and segmentation~\cite{dou2018unsupervised}. 
These methods aim at learning transformations that map two shifted image domains to a common feature space, so that the learned model that works for one domain can be applied to another.
The methods have demonstrated very promising performance for the target tasks~\cite{tzeng2017adversarial,dou2018unsupervised}.

Combining the advantages of both supervised learning-based density regression methods and unsupervised learning-based domain adaptation methods, 
This study proposes a novel, manual-annotating-free automatic cell counting method. The proposed method is evaluated on experimental immunofluorescent microscopic images of human embryonic stem cells (hESC).

\section{Methodology}
\label{sec:methodology}

\subsection{Background: Density Regression-Based Automatic Cell Counting}
\label{ssec:density}

The goal of density regression-based cell counting methods is to learn a density function $F$ 
that can be employed to estimate the cell density map for a given image~\cite{lempitsky2010,xie2018}.
For a given image $X\in \mathbb{R}^{M\times N}$ that includes $N_c$ cells, 
the corresponding density map $Y\in \mathbb{R}^{M\times N}$ can be considered as the superposition of a set of $N_c$ normalized 2D discrete Gaussian kernels that are placed at the centroids of the $N_c$ cells. 
Let $S = \{ ({s_{k_x}},{s_{k_y}})\in \mathbb{N}^2: k = 1,2, ..., N_c\}$ represent the cell centroid positions in $X$. 
Each pixel $Y_{i,j}$ on the density map $Y$ can be expressed as
\begin{equation}
Y_{i,j} = \sum_{k=1}^{N_c} G_\sigma (i-{s_{k_x}},j-{s_{k_y}}), \quad \forall \quad i \in M, \quad j \in N, \\
\label{eq:map}
\end{equation}

\noindent where $G_\sigma(n_x,n_y) = C \cdot e^{-\frac{n_x^2+n_y^2}{2\sigma^2}}\in \mathbb{R}^{(2K_G+1)\times (2K_G+1)}, n_x, n_y = -K_G, -K_G+1,..., K_G$, is a normalized 2D Gaussian kernel that satisfies $\sum_{n_x=-K_G}^{K_G}\sum_{n_y=-K_G}^{K_G} G_\sigma(n_x,n_y) =1$. Here, $\sigma^2$ is the isotropic covariance, $(2K_G+1)\times (2K_G+1)$ is the kernel size, and C is a normalization constant.

The density regression-based cell counting process includes three steps: 
(1) map an image into a feature map,
(2) estimate a cell density map from the feature map, and 
(3) integrate the estimated density map for cell counting.
In the first step, each pixel $X_{i,j}$ can be assumed to be associated with a real-valued feature vector $\phi(X_{i,j})\in \mathbb{R}^Z$. 
The feature map $P\in \mathbb{R}^{M\times N \times Z}$ of $X$ can be generated using specific feature extraction methods, such as the dense scale invariant feature transform (SIFT) descriptor~\cite{vedaldi2010vlfeat}, ordinary filter banks~\cite{fiaschi2012}, 
or codebook learning~\cite{sommer2011ilastik}.
In the second step, the estimated density $\hat{Y}_{i,j}$ of each pixel $X_{i,j}$ can be obtained by applying a pre-trained density regression function $F$ on the given $\phi(X_{i,j})$:
\begin{equation}
	\hat{Y}_{i,j} = F(\phi(X_{i,j});\Theta),
	\label{eq:estimatemap}
\end{equation}

\noindent where $\Theta$ is a parameter vector that determines the function $F$.
Finally, in the third step, the number of cells in $X$, $N_c$, can be counted by integrating the estimated density map $\hat{Y}$ over the image region:
\begin{equation}
	N_c \approx \hat{N_c} = \sum_{i=1}^{M}\sum_{j=1}^{N} \hat{Y}_{i,j}.
	\label{eq:counting}
\end{equation}

A key task in density regression-based cell counting methods is learning the function $F$ by use of training datasets.
The learning of $F$ and the related cell counting method proposed in this study are described below.

\subsection{The Proposed Automatic Cell Counting Framework}
\label{ssec:method}

The proposed density regression-based automatic cell counting method is implemented by use of both supervised and unsupervised learning, and employs both annotated synthetic images and unannotated experimental images.
Combining the advantages of both supervised learning-based density regression methods and unsupervised learning-based domain adaptation methods, 
the proposed method can learn a DRM by use of annotated synthetic images without the need of manually-annotated experimental images. A domain adaptation model (DAM) will be learned by use of both unannotated synthetic and experimental images.

\subsubsection{Overview of the Proposed Method}
\label{sssec:overview}

\begin{figure}
	\begin{center}
	    \includegraphics[width=\textwidth]{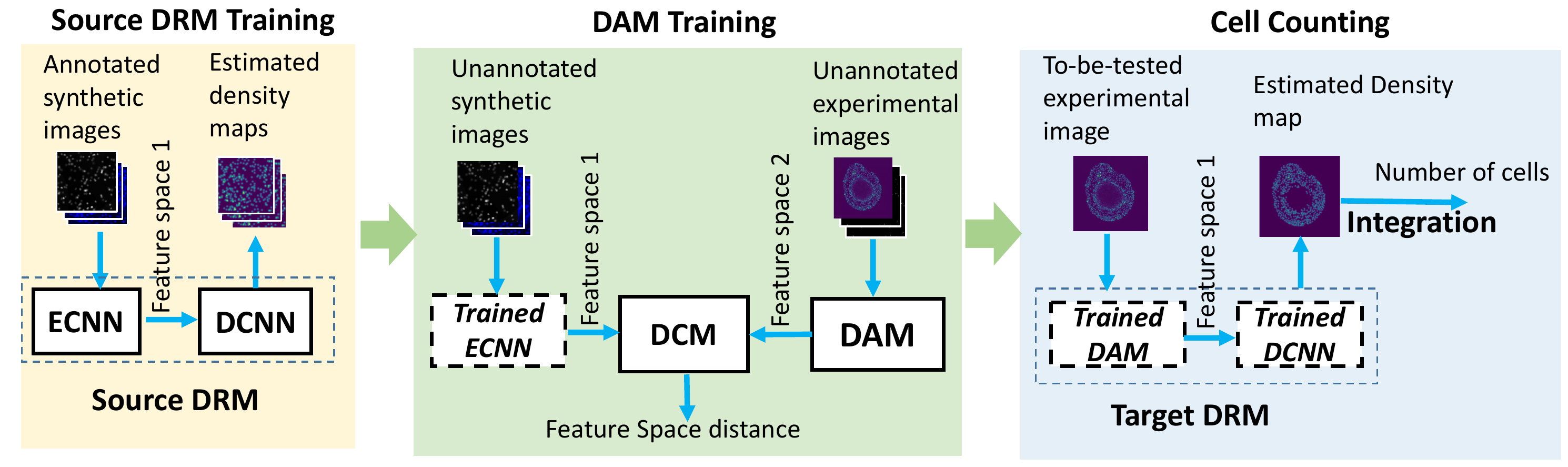}
	\end{center}
\caption{Overview of the proposed automatic cell counting framework.}
\label{fig:framework}
\end{figure}

The proposed method, shown in Figure~\ref{fig:framework}, has three phases: 1) Source DRM training (Section~\ref{sssec:drm}), 2) DAM training (Section~\ref{sssec:dam}), and 3) Density map estimation for cell counting based on the target DRM (Section~\ref{sssec:counting}).
Here, a source DRM represents the DRM trained with synthetic images (the source domain), while a target DRM is the domain-adapted DRM that can be employed for estimating the density map of a given experimental image (the target domain). 
In the source domain training phase, a source DRM consisting of an encoder CNN (ECNN) and decoder CNN (DCNN) is trained by use of supervised learning with a set of annotated images in the source domain. The trained ECNN maps the images in the source domain to a feature space of the source domain, while the DCNN maps the feature space to the corresponding density maps. The trained ECNN will be employed for training a DAM, while the trained DCNN is employed as part of the target DRM for density estimation and cell counting. In the DAM training phase, a DAM is trained in an unsupervised manner by use of the trained ECNN and a set of unannotated synthetic images and experimental images. The trained DAM will become part of the target DRM and will be employed to map the images in the target domain to a feature space that has minimum domain shift with that of the source domain. In the cell counting phase, the combination of the trained DAM and DCNN forms the target DRM to estimate the density map for a given experimental image.

%

\subsubsection{Source DRM Training}
\label{sssec:drm}

The first phase of the proposed method is to train a source DRM with annotated synthetic images, as shown in the left block of Figure~\ref{fig:drm}.
The trained source DRM then determines a density function $F$.
The source DRM is designed as a fully convolutional neural network (FCNN) that includes an encoder CNN (ECNN) and a decoder CNN (DCNN). This design is motivated by a network architecture described in the literature~\cite{xie2018}.
The ECNN encodes synthetic images to a low-dimensional and highly-representative feature space, while the DCNN decodes the feature space to estimate the corresponding density map. 
The architectures of the ECNN and DCNN are shown in Figure~\ref{fig:drm}.
Each block in the ECNN or DCNN includes a chain of layers.
Here, CONV represents a convolutional layer, Pool represents a max-pooling layer, and \lq\lq{}Up\rq\rq{} represents an up-sampling layer. 
There are a total of eight CONV layers in the ECNN and DCNN. 
The numbers of kernels in these eight CONV layers are set to 32, 64, 128, 512, 128, 64, 32, and 1, respectively.
The size of each kernel in the first seven CONV layers and that of the kernel in the last CONV layer are set to $3\times 3$ and $1\times 1$, respectively.

Let the source DRM be denoted as $F(X^s; \Theta)$, 
where $X^s$ represents a synthetic image in the source domain, 
and $\Theta = (\Theta^e, \Theta^d)$ represents the parameter vectors in the ECNN and the DCNN.
Training the source DRM is equivalent to learning a function $F$ with a parameter vectors $(\Theta^e, \Theta^d)$ that maps $X^s$ to an estimated density map, $\hat{Y}$, such that
%
\begin{equation}
Y \approx \hat{Y} = F(X^s; \Theta^e, \Theta^d).
\end{equation}

Let the source ECNN and DCNN be denoted as $\mathcal{A}^s(X^s;\Theta^{e})$ and $\mathcal{D}^s(\mathcal{A}^s(X^s;\Theta^{e});\Theta^{d})$, respectively.
Therefore, $F(X^s; \Theta) = \mathcal{D}^s(\mathcal{A}^s(X^s; \Theta^{e}); \Theta^{d})$.
The ECNN and the DCNN are jointly trained with a given set of $B$ training data 
$G^s = \{(X^s_i,Y^s_i)\}_{i = 1, ..., B}$, 
where $X^s_i$ is the $i$-th synthetic image and $Y^s_i$ is its ground truth density map. 
The training process is implemented through the minimization of a loss function $L(\Theta)$ that is defined as
\begin{equation}
L(\Theta) = \frac{1}{B}\sum_{i=1}^{B}\left\lVert Y_i^s - F(X^s_i;\Theta)\right\rVert ^2.
\label{eq:loss1}
\end{equation}

The numerical minimization of $L(\Theta)$ is performed via a momentum stochastic gradient descent (SGD) method~\cite{bottou2010large}.
A trained $F$ with the optimized parameters $\Theta^*= (\Theta^{e*}, \Theta^{d*})$ is obtained in this step.
The trained ECNN, $\mathcal{A}^s(X^s;\Theta^{e*})$, will be employed for training the DAM as described next.

\subsubsection{DAM Training}
\label{sssec:dam}

The second phase of the proposed method is to train a DAM by use of images from both the source and target domains and the trained ECNN, $\mathcal{A}^s(X^s;\Theta^{e*})$.
The trained DAM is employed to map images in the target domain to a feature space that has minimum domain shift with the feature space of images in the source domain.

\begin{figure} [!ht]
\begin{center}
	\begin{subfigure}{0.7\textwidth}
		\includegraphics[width=\textwidth]{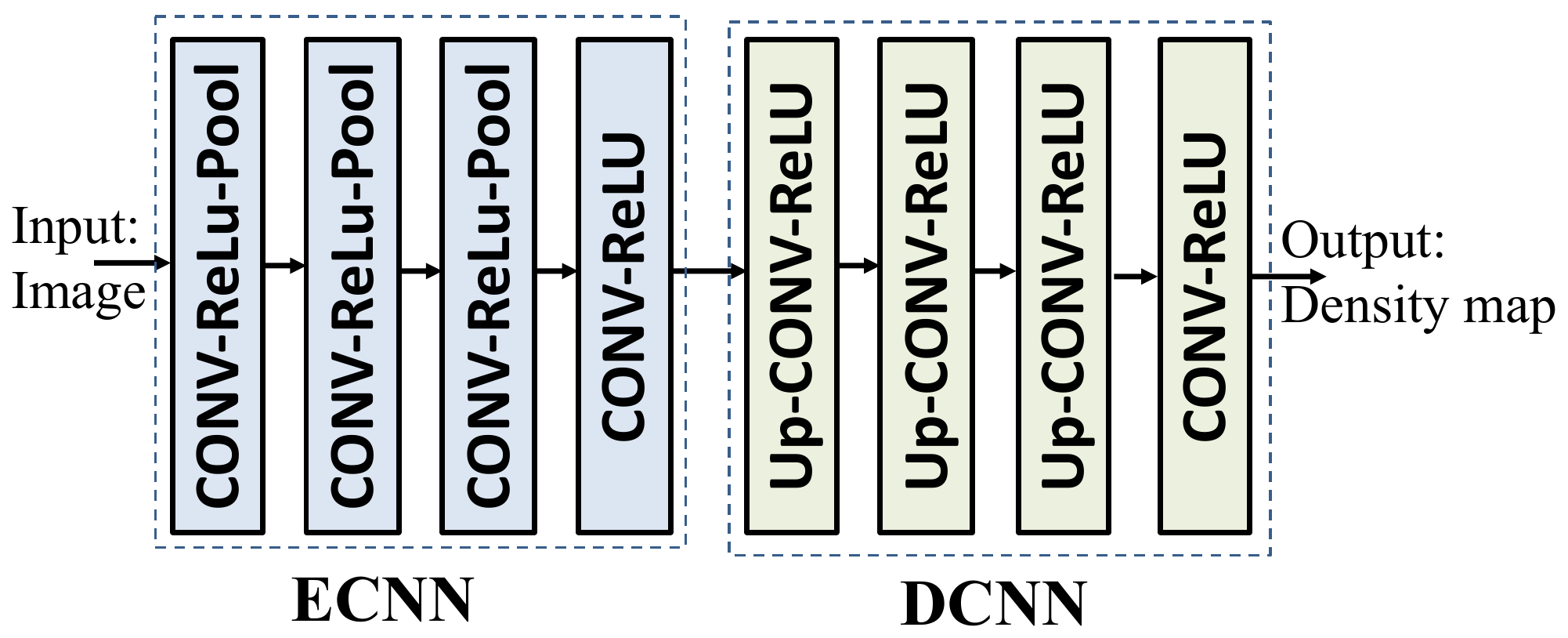}
		\caption{The architecture of the source DRM}
		\label{fig:drm}
	\end{subfigure}
   \break
	\begin{subfigure}{0.4\textwidth}
		\includegraphics[width=\textwidth]{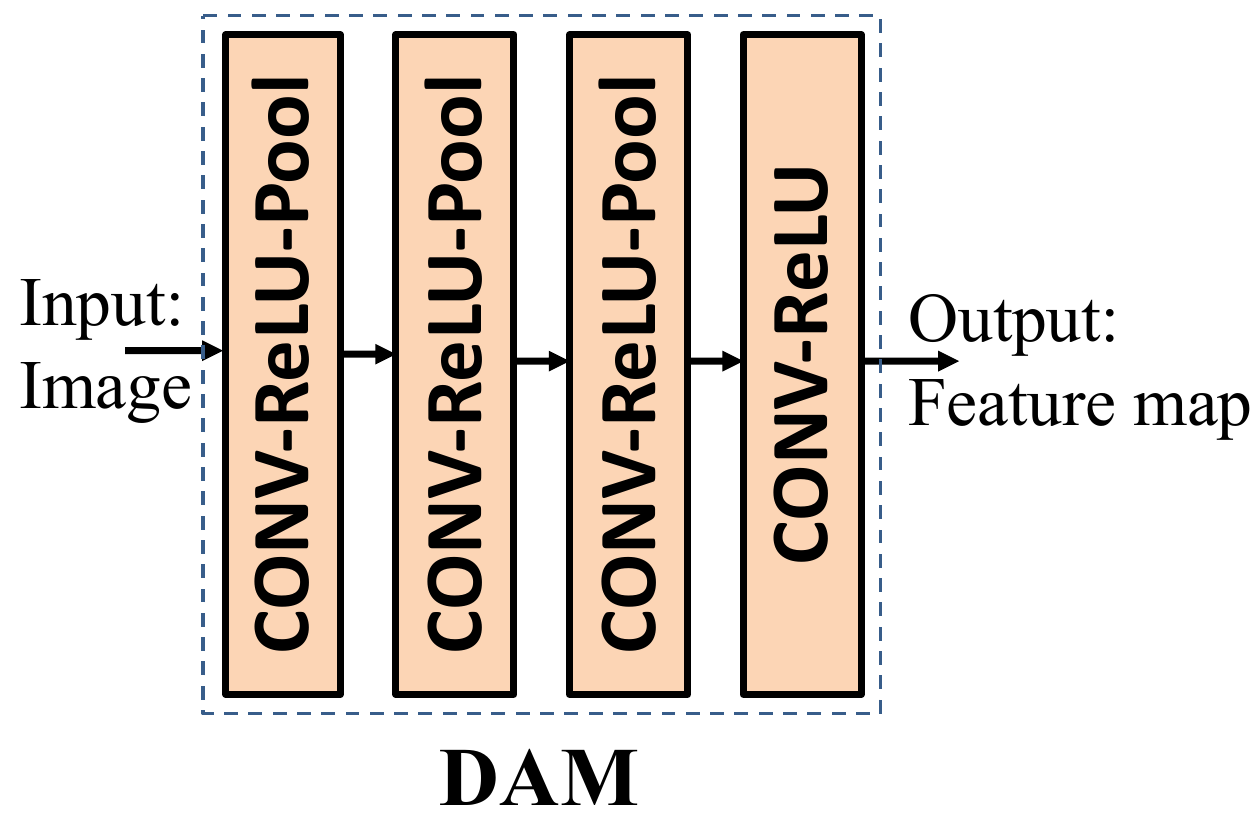}
    \caption{The architecture of the DAM}
    \label{fig:dam}
  \end{subfigure}
	\begin{subfigure}{0.55\textwidth}
		\includegraphics[width=\textwidth]{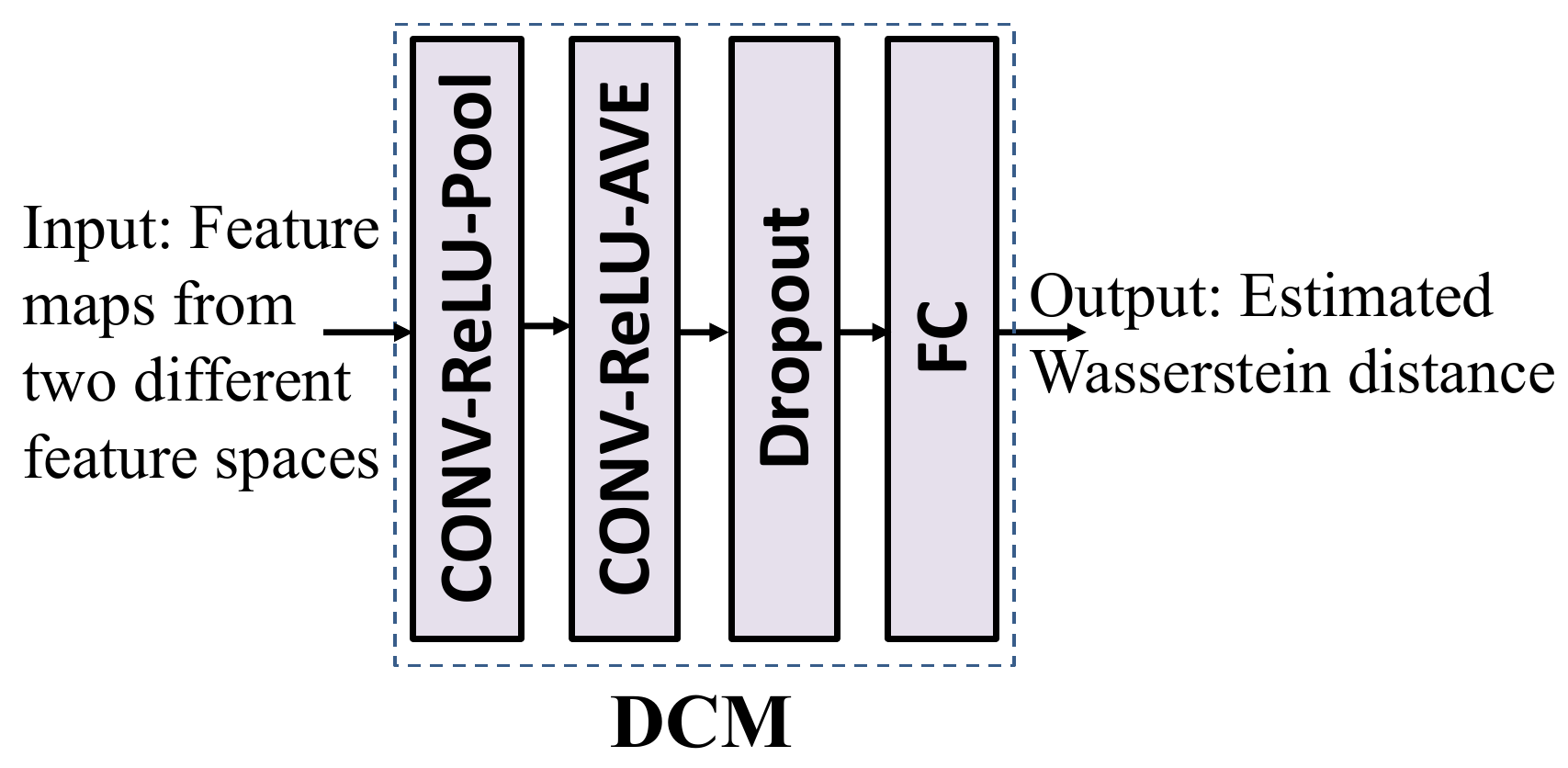}
    \caption{The architecture of the DCM}
    \label{fig:dcm}
  \end{subfigure}
\end{center}
\caption{
The network architectures of the source DRM (ECNN+DCNN), the DAM, and the DCM.
}
\label{fig:dam_training}
\end{figure}

The DAM training process is shown in the middle block of Figure~\ref{fig:framework}.
The network architecture of the DAM is shown in Figure~\ref{fig:dam}.
Let the DAM be denoted as $\mathcal{A}^t(X^t,\Theta^t)$.
The DAM is trained with unannotated source and target images through the minimization of domain shifts between the feature spaces of both the source and target domains. 
The Wasserstein distance is identified as a metric measuring this domain shift in this study. 
Due to the difficulty of directly computing the Wasserstein distance, as discussed in the literature~\cite{arjovsky2017wasserstein},
a domain critic model (DCM) is set up as a surrogate to estimate the Wasserstein distance for differentiating the feature distributions of the two domains.
The architecture of the DCM employed in this study is shown in Figure~\ref{fig:dcm}.
The DCM includes two CONV layers.
The numbers of kernels in these two CONV layers are set to 128 and 256, respectively. 
The size of each kernel is set to $3\times 3$. 
AVE represents an average pooling layer, and FC represents a fully connected layer. 
Dropout rate is set to 0.5, and the number of neurons in the FC layer is set to 256. 
The DAM and DCM are optimized iteratively via an adversarial loss function~\cite{arjovsky2017wasserstein} in an unsupervised manner by use of unannotated synthetic and experimental images.


\subsubsection{Density Estimation for Cell Counting}
\label{sssec:counting}
The final phase of the proposed method is density estimation for automatic cell counting on a given image
as shown in the right block of Figure~\ref{fig:framework}. 
The combination of the trained DAM and the trained DCNN forms a target DRM, 
$\mathcal{D}^s(\mathcal{A}^t(X^t;\Theta^{t*});\Theta^{d*})$.
The trained DAM maps an experimental image to a feature map in the feature space of the source domain,
and the trained DCNN estimates the density map of the image with this feature map. 
Finally, the number of cells is subsequently counted by integrating the estimated density map.

\section{Experimental Results}
\label{sec:experiment}

\subsection{Datasets}
\label{ssec:dataset}

The datasets used in this study are described in Table~\ref{tab:dataset}.
In this experiment, $200$ synthetic bacterial fluorescent microscopic cell images of $256\times 256$ pixels each were generated by use of methods described in the literature~\cite{lehmussola2007}.
The synthetic images were annotated automatically when generated. The ground truth density map of each synthetic image was generated by placing normalized Gaussian kernels at each annotated cell centroid in the image, according to the methods introduced in Section~\ref{ssec:density}. The values of $\sigma$ and $K_G$ were set to $3$ pixels and $10$ pixels, respectively.

In addition, $122$ experimental immunofluorescent microscopic hESC images of $512\times 512$ pixels each were employed.
In 10 of these $122$ images, the centroids of each cell were manually annotated.
The density map for each of the $10$ images was generated with $\sigma$ and $K_G$ being $3$ pixels and $10$ pixels, respectively.
These density maps were employed as the ground truth to evaluate the cell counting performance.
%
\begin{table}[ht]
\caption{Datasets employed in this study}
\begin{center}
\begin{tabular}{c c c c}
\hline\hline
\textbf{Images}		& \textbf{Annotated synthetic}	& \textbf{Unannotated hESC}& \textbf{Annotated hESC}  \\
\textbf{Dataset size}	& 200 						& 112 						& 10 \\
\textbf{Image size (pixels)} & $256\times 256 $ 		& $512 \times 512$ 			&  $512\times 512$\\
\textbf{Purpose}		& DRM, DAM training			& DAM training       			& Cell counting evaluation \\
\hline\hline 
\end{tabular}
\end{center}
\label{tab:dataset}
\end{table}

Each pair of a synthetic image and a ground truth density map was employed for the supervised learning of the source DRM.
Out of the $200$ samples, $160$ were employed for training the source DRM, and the remaining $40$ were employed as the validation set.
Different from the source DRM training, unannotated synthetic and experimental data were employed in the DAM training phase.
The same $200$ synthetic cell images (without annotation) and $112$ unannotated experimental hESC images
were employed as the source and target image sets respectively, 
and were used for the adversarial learning of the DAM and DCM. 
The DAM learning process was described in Section~\ref{sssec:dam} above. Ten manually-annotated images and their density maps were employed as the ground truths to evaluate the cell counting performance of the proposed method.

\subsection{Method Implementation}
\label{ssec:implement}

The training and evaluation of the proposed method were performed on a NVIDIA Titan X GPU with 12GB of VRAM. 
Software packages employed in our experiments included Python 3.5, Keras 2.0, and Tensorflow 1.0.
In the source DRM training, the learning rate was set to $0.001$ and the batch size was set to $100$. The learning rate is a hyper-parameter that controls the stride of updating the values of the parameters in a to-be-trained model in each iteration.
A mean square error (MSE) loss function was employed for model training.
After $3000$ epochs, the model that resulted in the lowest MSE in the validation set was stored as the to-be-employed source DRM.

In the DAM training, the DAM was initialized with the weights of the trained ECNN.
The learning rates for training the DAM and DCM were both set to $10^{-8}$. 
In each iteration, $100$ images were randomly selected from the source image set, 
and another $100$ images were randomly selected from the target image set. 
Each selected target image was randomly cropped into an image of $256 \times 256$ pixels,
so that the inputs of the ECNN and those of the DAM had the same sizes.
The DAM and DCM were iteratively optimized via an adversarial loss~\cite{arjovsky2017wasserstein}.


\subsection{Results}
\label{ssec:result}

We compared the results obtained from the proposed method (denoted as Adaptation) with those from two other methods.
The first alternative method applied the source DRM directly to the experimental cell images (denoted as Source-only).
The second method was a fully convolutional regression network (FCRN)-based DRM~\cite{xie2018} that was trained with annotated experimental images (denoted as Annotated-train).
For the Annotated-train method, $5$ out of the $10$ annotated images were employed 
to train the FCRN-based DRM, and the remaining $5$ were employed for the model validation.

The performances of the three cell counting methods were measured by the mean of absolute errors (MAE) and standard deviation of absolute errors (SAE).
MAE measures the mean of absolute errors between the estimated cell counts and their ground truths for all 10 annotated hESC images, while SAE measures the standard deviation of the absolute errors.
The results of evaluating the three methods on all 10 images are shown in Table~\ref{tab:mae}.
In terms of MAE and SAE, the proposed method demonstrates superior cell counting performance to the Source-only and Annotated-train methods.
\begin{table}[ht]
\caption{Performance of the proposed cell counting method} 
\begin{center}       
\begin{tabular}{c c c c} 
\hline\hline
Performance  	& Adaptation 		& Source-only 			& Annotated-train\\
MAE $\pm$ SAE 	&  \textbf{35.82$\pm$24.98}	& 170.06 $\pm$ 50.58 	& 39.84 $\pm$ 25.24\\
\hline\hline
\end{tabular}
\end{center}
\label{tab:mae}
\end{table}
\begin{figure}
	\begin{center}
	    \includegraphics[width=\textwidth]{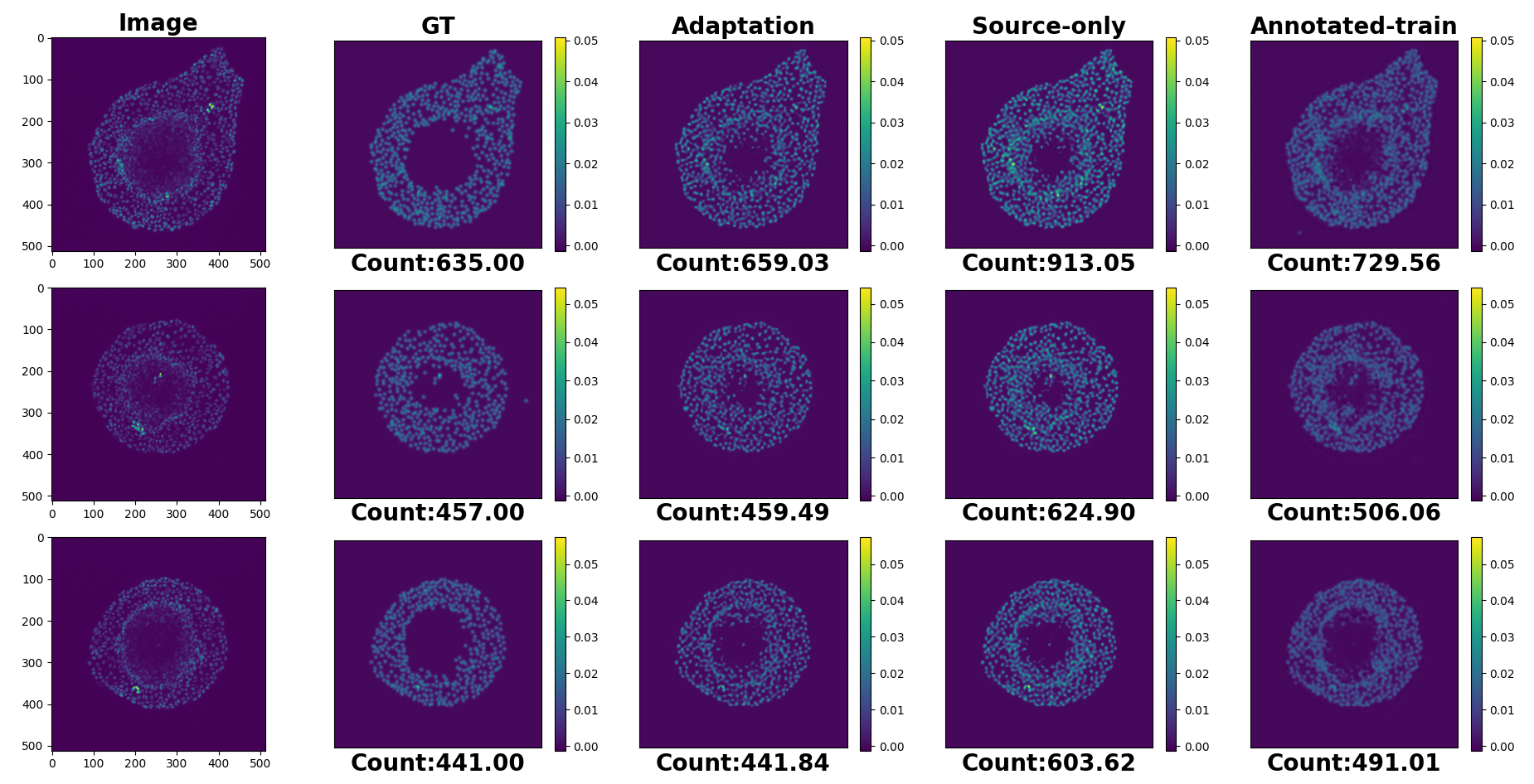}
	\end{center}
\caption{Density map estimation for 3 out of the 10 annotated hESC image examples.}
\label{fig:density_results}
\end{figure}

Additionally, the density maps of three testing hESC images estimated by use of the three methods are shown in Figure~\ref{fig:density_results}.
The figures in each row display the results corresponding to different images.
In each row,  the sub-figures from left to right show the experimental image, the ground truth density map, 
and the density maps estimated with the proposed Adaptation method, the Source-only method, 
and the Annotated-train method, respectively.
The ground truth number of cells and the number counted from the estimated density maps are indicated at the bottom of each sub-figure.
As shown in Figure~\ref{fig:density_results}, the proposed method can estimate a density map that is visually similar to the ground truth density map. The density maps estimated by use of the Source-only method is much denser than the ground truth, while those estimated by the Annotated-train method are blurred.

%


\section{Discussion}
\label{sec:discussion}
In this work, a domain adaptation-based density regression framework is proposed for automatic microscopic cell counting.
Instead of training a DRM using the experimental images that require manual annotating, in this study, we propose to 1) train a source DRM with annotated synthetic images and 2) train a DAM 
with both unannotated synthetic and experimental images to minimize the domain shift between the source and target domains.
The trained source DRM and DAM are jointly employed as a target DRM for automatic cell counting in experimental images. 
The proposed method reduces or even eliminates the need to manually annotate experimental images and can greatly improve the generalization of the DRM trained with annotated synthetic images. 
To the best of our knowledge, this is the first study in which a domain adaptation method is employed to support the task of automatic cell counting in microscopic images.

In addition, the proposed framework allows many flexible modifications. 
For example, the FCNN network employed in the training of the source DRM can be 
replaced with other FCNN networks when different tasks need to be addressed.
The choice of the FCNN in the current study is motivated by the success of deep neural networks in other computer vision tasks, including image classification~\cite{he2016}, segmentation~\cite{ronneberger2015, he2018}, and object detection~\cite{ren2015}.
Although the network architecture of the FCNN is fixed in this study, tuning of the network architecture is still an open question in deep learning-based researches, but it is not the focus of this work.



\section{Conclusion}
\label{sec:conclusion}

A novel, manual-annotating-free density regression framework is proposed for automatic microscopic cell counting. 
The proposed method integrates a supervised learned DRM and an unsupervised learned DAM for automatic cell counting, and achieves promising cell counting performance.

\acknowledgments 
 
This work was supported in part by award NIH R01EB020604, R01EB023045, R01NS102213, and R21CA223799.  

\bibliography{references} 
\bibliographystyle{spiebib} 

\end{document}